\documentclass[letterpaper, 10 pt, conference]{ieeeconf}  

\IEEEoverridecommandlockouts                              

\title{\LARGE \bf
Inline Photometrically Calibrated Hybrid Visual SLAM
}
\usepackage{cite}
\usepackage{amsmath,amssymb,amsfonts}
\usepackage{algorithmic}
\usepackage{graphicx}
\usepackage{textcomp}
\usepackage[export]{adjustbox}
\usepackage{xcolor}
\usepackage{url}
\usepackage{hyperref}
\usepackage{booktabs}
\usepackage{multirow}
\usepackage{tabularx}
\usepackage{adjustbox}
\usepackage{graphicx}
\usepackage[absolute,overlay]{textpos}

\author{
        Nicolas Abboud, Malak Sayour, Imad H. Elhajj, John Zelek, Daniel Asmar \\
\thanks{N. Abboud, M. Sayour, I. H. Elhajj, and D. Asmar are with the Vision and Robotics Lab, Maroun Semaan Faculty of Engineering and Architecture, American University of Beirut, 1107 2020, Riad El Solh, Beirut,
Lebanon; email: nfa53@mail.aub.edu, ms423@aub.edu.lb, ie05@aub.edu.lb, da20@aub.edu.lb.}
\thanks{J. Zelek is with the department of System Design Engineering at the University of Waterloo, 200 University Avenue West, Waterloo, Ontario, Canada; email: jzelek@uwaterloo.ca.
}
}

\usepackage{xspace}
\makeatletter
\DeclareRobustCommand\onedot{\futurelet\@let@token\@onedot}
\def\@onedot{\ifx\@let@token.\else.\null\fi\xspace}

\def\etal{\emph{et al}\onedot}

\makeatother

\begin{document}
\maketitle

\begin{abstract}
This paper presents an integrated approach to Visual SLAM, merging online sequential photometric calibration within a Hybrid direct-indirect visual SLAM (H-SLAM).  Photometric calibration helps normalize pixel intensity values under different lighting conditions, and thereby improves the direct component of our H-SLAM. A tangential benefit also results to the indirect component of H-SLAM given that the detected features are more stable across variable lighting conditions. Our proposed photometrically calibrated H-SLAM is tested on several datasets, including the TUM monoVO as well as on a dataset we created. Calibrated H-SLAM outperforms other state of the art direct, indirect, and hybrid Visual SLAM systems in all the experiments. Furthermore, in online SLAM tested at our site, it also significantly outperformed the other SLAM Systems.
\end{abstract}

\section{Introduction}
The advent of Visual Simultaneous Localization and Mapping (V-SLAM) has been a cornerstone in realizing autonomous navigation and perception systems. Advancements in V-SLAM have been realized with the development of hybrid V-SLAM systems that merge direct and indirect methods, and leverage the strengths of both methods to overcome their inherent limitations. Direct methods \cite{engel2017direct}, which estimate motion and structure directly from pixel intensity variations are highly susceptible to variations in illumination, often resulting in decreased robustness and accuracy in dynamically lit environments \cite{bergmann2017online}. The integration of indirect methods \cite{mur2015orb} (which rely on feature extraction and matching) with direct methods offers a balanced solution, enhancing the system's adaptability and reliability across a wide range of scenarios by ensuring stable feature tracking even under challenging lighting conditions.

\begin{figure}[t]
  \centering
  \includegraphics[width=\linewidth]{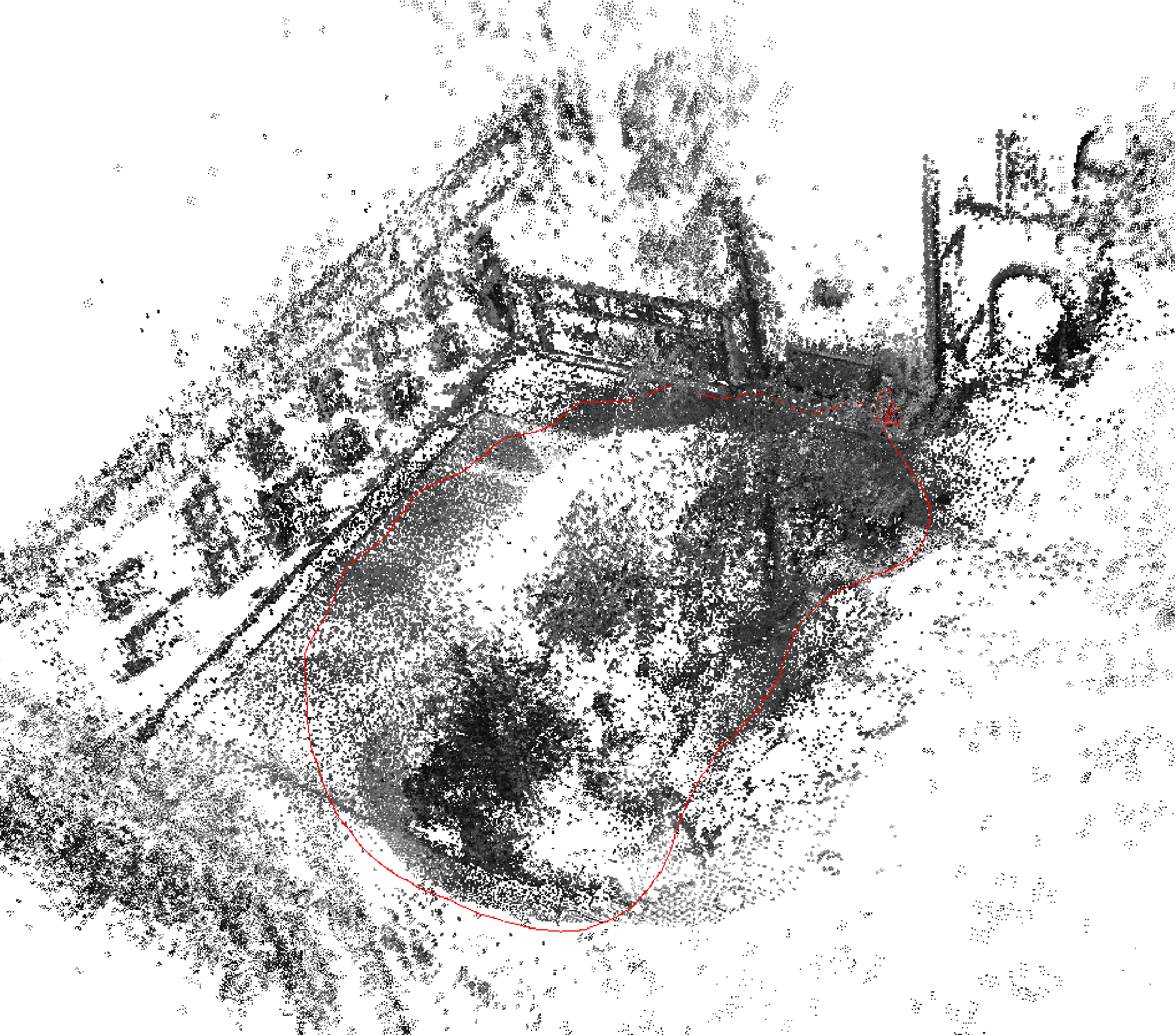}
  \caption{Sample output of the photmetrically calibrated hybrid SLAM system run on Sequence 30 of the TUM monoVO dataset.}
  \label{hslamresult}
\end{figure}

Photometric calibration plays a crucial role in further bolstering the capabilities of the direct component within hybrid SLAM systems (Fig.\ref{hslamresult}). By accurately mapping the scene's radiance to the camera's intensity values, photometric calibration addresses critical aspects such as radiometric calibration and vignetting compensation.
Radiometric calibration, is required to estimate the Camera Response Function (CRF), thereby ensuring the authenticity of intensity values relative to real-world irradiance\cite{grossberg2003determining}. This is particularly vital in enhancing system robustness under fluctuating illumination conditions \cite{bergmann2017online}. Vignette compensation, conversely, rectifies the radial intensity reduction towards the image periphery, a common lens-induced aberration, thereby achieving uniform sensitivity across the camera's field of view. This uniformity is critical for improving the quality and reliability of feature detection and tracking in visual odometry.

Our proposed system integrates our Online Sequential Photometric Calibration (OSPC) technique \cite{haidar2023ospc} within our Hybrid direct-indirect visual SLAM (H-SLAM) system \cite{younes2023h}, utilizing exposure values from frame metadata to sequentially estimate the camera response function and vignette. The estimated photometric parameters are used to rectify the input frames to the V-SLAM system; thus improving the visual odometry module's adaptability within the hybrid SLAM architecture, and providing a robust and consistent basis for feature tracking and depth estimation. This integration enriches the OSPC component's effectiveness, which depends on precise feature extraction to identify corresponding points across frames. As a result, the enhanced feature extraction feeds back into the OSPC, leading to more accurate photometric calibration. This synergy creates a cycle within the system, where improved photometric calibration further bolsters the visual odometry component, thereby strengthening the entire SLAM framework's performance and reliability.

The practicality and robustness of this advancement have been validated through both dataset and real-world testing, diverging from the conventional reliance solely on dataset evaluations. The real-world focus offers a comprehensive understanding of the system's performance in dynamic and unpredictable environments.

This paper presents a real-time Visual SLAM system, that merges OSPC's photometric estimation with H-SLAM's efficient methodology. The source code is available at \href{https://github.com/AUBVRL/HSLAM_docker/tree/hslam_ospc}{this link}. The key contributions are as follows:

\begin{enumerate}
\item \textbf{Unified Hybrid Direct-Indirect SLAM}: we leverage the strengths of both direct and indirect approaches to achieve robust camera pose estimation, demonstrating accuracy on par with or surpassing state-of-the-art monocular SLAM/Odometry system. This system efficiently computes both local and global map representations in a joint representation, enhancing the re-utilization of global map points and reducing memory consumption significantly compared to traditional methods.
\item \textbf{Adaptive in-Line photometric calibration}: building upon OSPC's approach to photometric calibration, our system implements this technique adaptively and in-line within a SLAM pipeline, directly contributing to the overall improvement of hybrid V-SLAM performance.
\item \textbf{Enhanced accuracy and robustness in feature-deprived environments}: our approach achieves enhancements in localization and mapping precision by uniquely keeping track of both pose-pose and co-visibility  constraints, advancing robustness in feature deprived environments, a common challenge in real-world applications.

\item \textbf{Comprehensive experimental validation}: experiments and evaluations conducted in real-world scenarios validate the effectiveness and robustness of our proposed system. These tests highlight the system's adaptability to various conditions and its practical applicability.
\end{enumerate}

\section{Related Work}
In this section we review the existing literature pertinent to our study, highlighting advancements and identifying gaps that our research aims to address.
\subsection{Hybrid Visual SLAM: Bridging Direct and Indirect Methods}
Feature-based methods involve the extraction and matching of a limited set of keypoints, aiming to minimize reprojection errors for pose estimation. This approach, as exemplified by the works of \cite{geiger2011stereoscan},\cite{klein2007parallel}, \cite{mur2015orb} and \cite{pire2015stereo} relies on the robustness of modern feature extraction algorithms to handle variations in image intensity and geometric noise. However, it should be noted that their performance tends to deteriorate in environments lacking distinctive textures or when images are affected by photometric noise, as discussed by \cite{engel2017direct}.
Direct methods as introduced by \cite{engel2017direct}, \cite{engel2013semi}  and \cite{newcombe2011dtam} do not rely on feature extraction. Instead, they estimate motion and structure by minimizing photometric errors directly, which represent discrepancies in image intensity. Nevertheless, it is important to acknowledge that these methods are highly sensitive to changes in image intensity caused by factors such as varying lighting conditions and camera exposure times \cite{engel2017direct}.

Hybrid methods attempt to combine the strengths of both feature-based and direct methods. In the work of Lee and Civera \cite{lee2018loosely}, a loosely-coupled approach integrates a feature-based method and a direct method, allowing one to support the other when faced with challenges. However, this architecture, demands additional computational resources as it entails maintaining both methods simultaneously. Semi-direct Visual Odometry (SVO) \cite{forster2016svo}, represents an efficient hybrid system that combines direct image alignment using initially detected features with Bundle Adjustment (BA) based on reprojection error for pose estimation refinement but lacks back-end optimization and re-localization. Other researchers opt to first optimize the reprojection error to obtain an initial coarse camera pose and then refine the result using a direct image alignment algorithm, as demonstrated by \cite{kim2019autonomous}, \cite{krombach2017combining}, and \cite{luo2022hybrid} . This approach can be unified, as shown by Younes \etal \cite{younes2019unified}, to yield a feature-assisted visual odometry model.

Recent research into this unified approach addresses the intricate challenges encountered by V-SLAM systems in representing local and global maps, primarily stemming from parametrization disparities between direct and indirect data sources \cite{klein2007parallel} \cite{mur2015orb}. Indirect methods rely on multi-view geometry to construct point clouds (X, Y, Z), while direct methods hinge on small frame-to-frame motion for inverse depth parametrization \cite{civera2008inverse}. Additionally, this unified approach confronts the significant challenge faced by hybrid methods \cite{lee2018loosely}, \cite{luo2022hybrid}, and \cite{younes2019unified} which is the non-interchangeability of triangulation techniques. H-SLAM\cite{younes2023h} effectively surmounts these issues by adopting a versatile map representation that seamlessly adapts to function as either an (X, Y, Z) point cloud or an inverse depth parametrization, offering a comprehensive solution to the complexities faced by V-SLAM systems and hybrid approaches.
\subsection{Photometric Calibration: Enhancing Image Understanding}
In the domain of radiometric calibration for visual algorithms, a central concern revolves around the consistent maintenance of intensity across frames. Conventional approaches, as outlined by \cite{mann1995being } and \cite{mitsunaga1999radiometric}, typically rely on assumptions about the camera's response function. Unfortunately, these assumptions can lead to inaccuracies due to non-monotonic response functions \cite{grossberg2003determining}. To tackle this problem effectively, Grossberg and Nayar introduced the Empirical Model of Response \cite{grossberg2003space}, which operates on single images but does not recover the vignette or the exposure times. Moreover, authors of \cite{lin2004radiometric} and \cite{lin2005determining} developed CRF recovery methods using color and grayscale imagery, but they do not account for vignetting and exposure time variations. Multi-image static scene methods, like those developed by \cite{debevec2023recovering}, \cite{mitsunaga1999radiometric}, and \cite{engel2016photometrically}, are not suitable for dynamic video sequences due to their reliance on multiple exposures of a static scene. Zheng et. al \cite{zheng2008single} proposed a vignette recovery approach, but they also neglect exposure time estimation and assume a pre-calculated CRF. Building on this foundation, Kim et. al \cite{kim2008robust} enhanced the process by optimizing exposure differences and optical flow simultaneously, thereby improving robustness in the face of changing image intensity. However, their approach demands a significant number of center-symmetric samples and utilizes a joint estimation method, limiting its real-time feasibility. More recently, Bergmann et. al \cite{bergmann2017online} proposed an online photometric calibration system to recover radiometric parameters from auto-exposure videos. However, their method also employs joint estimation, requiring a point from the camera response function's ground truth data to resolve ambiguities. An ambiguity-free estimation for the camera response function was proposed by \cite{mo2019ambiguity}, but it ignores vignetting effects. We introduce  \cite{haidar2023ospc} a novel photometric calibration method using sequential optimization to independently estimate the camera response function, vignette, and exposure values, thereby avoiding the ambiguities and computational complexities typical of joint optimization methods. 

\subsection{Photometrically Calibrated Hybrid Visual SLAM}
Numerous recent endeavors have aimed to integrate photometric calibration techniques into V-SLAM systems to optimize the visual odometry module. In their work, liu \etal \cite{liu2021optical} introduce a real-time photometrically calibrated monocular direct SLAM system, showcasing significant advancements; however, it does exhibit certain limitations. Specifically, this system solely relies on a direct sparse method, foregoing the versatility offered by hybrid approaches. Furthermore, the real-time photometric calibration technique proposed introduces computational complexity through a joint optimization approach, which potentially affects real-time performance. The system's effectiveness is significantly influenced by lighting conditions, possibly leading to error accumulation and impacting its long-term reliability in various operational scenarios.

Conversely, Luo \etal \cite{luo2022hybrid} introduce a Hybrid Sparse Odometry system , integrating feature-based and direct methods V-SLAM. Their system aims to enhance robustness against variations in image intensity and motion blur. It incorporates photometric calibration into the visual odometry component, but this process is also performed as a joint optimization, introducing complexity and ambiguities in exposure estimation that may not be resolved without ground truth data. This poses potential challenges to the system's accuracy and reliability, especially in real-world applications where processing speed is paramount. Furthermore, the study lacks an extensive comparison of their approach with existing methods, particularly in real-world diverse scenarios, leaving its practical effectiveness and adaptability to various conditions and environments relatively unproven.
\begin{figure*}[ht]
      \centering
      \includegraphics[width=\linewidth, height=0.55\linewidth]{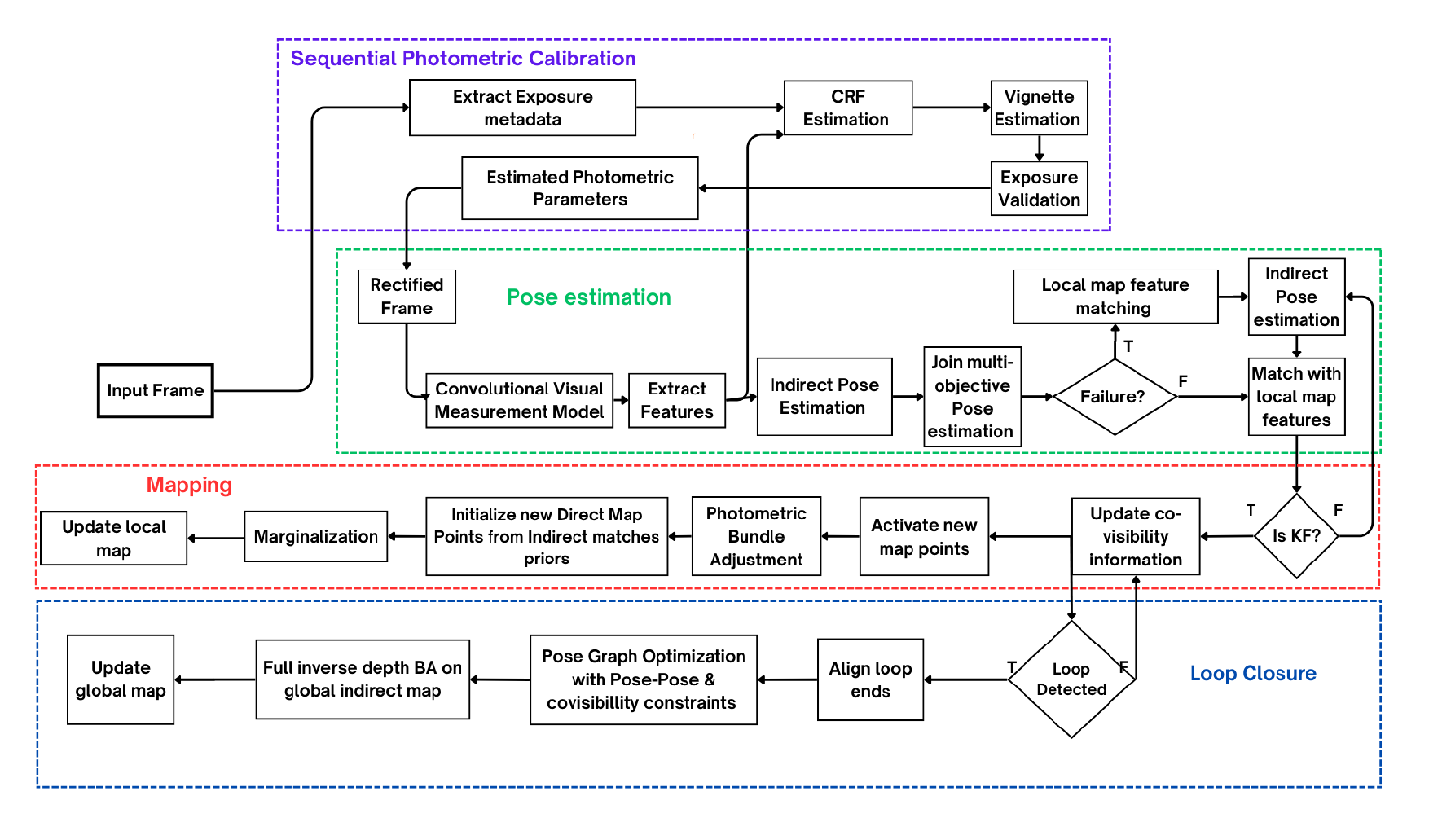}
      \caption{Diagram of the proposed system showing the integration of sequential photometric calibration with the multi-threaded architecture of Hybrid SLAM}
      \label{sys-diag}
\end{figure*}
In response to these limitations, we propose an innovative solution that marries a sequential photometric calibration method with a hybrid direct-indirect SLAM system, effectively addressing the shortcomings identified in prior research. Our approach simplifies the calibration process, reducing computational complexity while preserving accuracy. Rigorous testing across a range of operational scenarios underscores the system's reliability and precision in real-world environments. In summary, our hybrid SLAM system, enriched with streamlined photometric calibration, presents an enhancement in performance, accuracy, and adaptability in V-SLAM systems.

\section{Proposed System}
This section presents a framework integrating the OSPC within the H-SLAM architecture, designed to enhance the robustness, accuracy, and adaptability in SLAM applications. The integrated system coalesces the strengths of OSPC's sequential photometric calibration with H-SLAM's advanced feature tracking, pose estimation, and map maintenance capabilities ensuring consistency and coherency of the spatial mapping.

Here, we detail the technical nuances of the proposed integrated framework, emphasizing the synergistic interplay between OSPC and H-SLAM components.

\subsection{Descriptor sharing and feature extraction interplay}
The ORB feature extraction utilized in H-SLAM plays an important role in ensuring feature consistency, which is foundational for OSPC's process. The descriptor sharing inherent in H-SLAM enhances OSPC by providing a rich, multi-faceted representation of each feature, which is necessary for the tracking and photometric rectification processes employed.

The robustness of ORB features, combined with the multi-descriptor approach, ensures that OSPC tracks features consistently across frames, a vital prerequisite for accurate photometric rectification. The rich data provided by descriptor sharing empowers OSPC's calibration algorithms, enabling them to accurately estimate photometric parameters, thereby refining the photometric model used across the system.

\subsection{Sequential in-line photometric calibration}
The OSPC's sequential photometric calibration operates as an in-line process within the H-SLAM architecture. The calibration is sequential and operates until a pre-defined validation threshold is reached, after which the estimated photometric parameters are set. Subsequent frames undergo photometric rectification based on these set parameters, maintaining consistency in the visual information processed by H-SLAM.

\textbf{Integration into H-SLAM's multi-threaded architecture:} OSPC is seamlessly integrated as a fourth thread in H-SLAM's multi-threaded architecture, originally consisting of dedicated threads for mapping, pose estimation, and loop closure (Fig. \ref{sys-diag}). This integration ensures that the sequential photometric calibration operates concurrently with H-SLAM's other processes without impeding performance. The decoupled nature of this architecture allows OSPC to perform photometric rectification on each frame independently, contributing to the overall robustness and adaptability of the system without imposing additional computational burden on the core SLAM processes.

\textbf{Mathematical formulations:}
the photometric calibration in OSPC optimizes photometric parameters as follows:  
\begin{equation}
\frac{f^{-1}(M_1)}{f^{-1}(M_2)} = \frac{e_1}{e_2},
\end{equation} representing the irradiance ratio between pairs of frames. $f^{-1}$ represents the inverse camera response function, $M_1$ and $M_2$ are the corresponding intensity values, and $e_1$ and $e_2$ are the exposure value pairs. Similarly, the vignetting effect ($V$) is estimated by incorporating radial movement ($R_1$, $R_2$) alongside the CRF formulation:
\begin{equation}
\frac{f^{-1}(M_1)}{f^{-1}(M_2)} = \frac{e_1}{e_2} \frac{V(R_1)}{V(R_2)}         ,
\end{equation}Exposure validation and CRF estimation are as follows:
\begin{equation}
{k} = \frac{1}{N} \sum \frac{f^{-1}(M_1)}{f^{-1}(M_2)} \frac{V(R_2)}{V(R_1)} ,
\end{equation}
in which $N$ stands for the total number of corresponding points per image pair. These equations are fundamental to OSPC and enable sequential optimization and validation of photometric parameters, ensuring accurate calibration within the multi-threaded H-SLAM framework without overburdening its core processes.

\subsection{Harmonized integration: joint optimization, map maintenance and loop closure enhancement}
The convergence of OSPC's photometric rectification with H-SLAM's joint optimization, hybrid connectivity graphs, and loop closure synergistically enhances the system's performance. It ensures not only geometric coherency and photometric accuracy but also computational efficiency in the loop closure process.

\textbf{Optimized joint multi-Objective pose estimation}: photometric rectification, facilitated by OSPC, plays an important role in minimizing residuals in H-SLAM's joint multi-objective pose estimation. By rectifying each frame, OSPC ensures that the input to H-SLAM's pose estimation module is void of photometric distortions, leading to a significant reduction in both geometric and photometric residuals. The photometric residuals are calculated using the same method as DSO \cite{engel2017direct}, and the geometric residuals are the difference between the predicted and perceived key-point positions. The joint multi-objective pose optimization in H-SLAM is defined as an optimization problem:
\begin{equation}
        \text{argmin } e(\xi) = \text{argmin } \left( \frac{\lVert e_p(\xi) \rVert_{H_\delta}}{n_p \sigma_p^2} + K \frac{\lVert e_g(\xi) \rVert_{H_\delta}}{n_g \sigma_g^2} \right) ,
\end{equation}
where $\lVert \cdot \rVert_{H_\delta}$ denotes the Huber norm, $e_p(\xi)$ and $e_g(\xi)$ are the photometric and geometric residuals respectively, balanced by the number of features $n_p, n_g$, their variances $\sigma_p^2, \sigma_g^2$, and a utility function $K$ defined as:
\begin{equation}
    K = \frac{5e^{-2l}}{1 + e^{\frac{30-N_g}{4}}} ,
\end{equation}
where $l$ is the pyramid level and $N_g$ is the number of current inlier geometric matches, dynamically adjusts the optimization's focus, prioritizing geometric residuals in early stages and shifting towards photometric residuals as optimization progresses.
    
This approach ensures that the input to H-SLAM's pose estimation module is rigorously optimized for both geometric coherence and photometric consistency, translating into more coherent and precise map updates, enhancing the overall quality and reliability of the spatial map maintained by H-SLAM.

\textbf{Accurate hybrid connectivity graphs and loop closure enhancement:}
the reduced residuals and subsequent enhancements in mapping data ensure that the co-visibility and pose-pose constraints used in the hybrid connectivity graphs are based on precise data, enhancing the overall quality and utility of these graphs in the SLAM process. Precise co-visibility and pose-pose constraints in the hybrid connectivity graphs, facilitate accurate loop alignment and coherent loop closure. The accurate loop closure, aided by the consistent photometric information and the robust connectivity graphs, effectively minimizes drift over time. This contributes to the overall integrity and accuracy of the global map, ensuring that the system maintains a high level of performance even in extensive operational environments.
    
\textbf{Impact on bundle adjustment:} with enhanced feature matching and tracking, the input to the bundle adjustment process (\textit{i.e.}, the correspondences between 2D features and 3D points) is more accurate, improving the precision of the 3D structure reconstructed by the inverse depth bundle adjustment. This makes the global mapping more robust and the pose estimation more reliable, even in dynamically changing environments.

By leveraging the synergy between photometric rectification, feature extraction, and joint multi-objective pose estimation the system achieves high levels of accuracy and robustness, while avoiding high computational costs.
\section{Experiments}
We validate our system through experiments conducted on both publicly available datasets and real-life scenarios. The experiments are designed to assess our system's performance under various conditions, comparing outcomes with and without our suggested photometric calibration approach, and benchmarking against established SLAM systems.
\begin{figure*}[ht]
      \centering
      \includegraphics[width=0.9\linewidth, height=70mm]{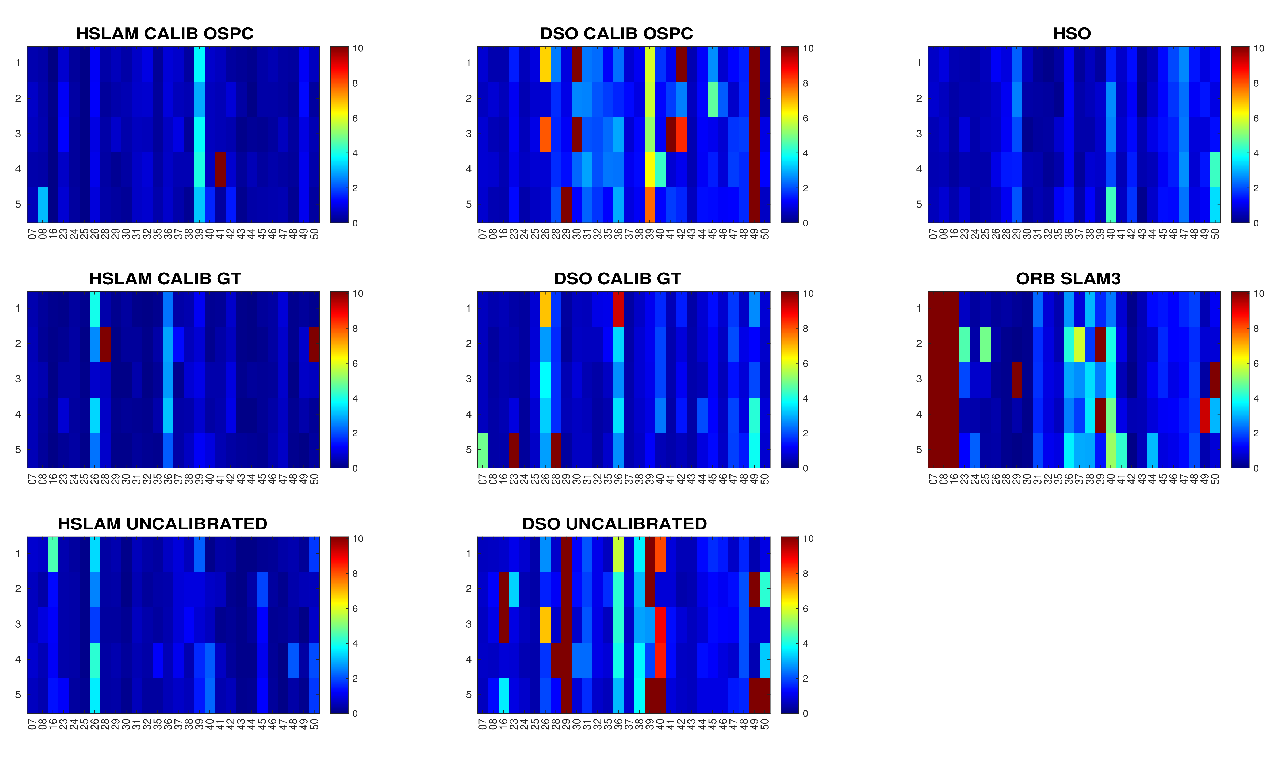}
      \caption{Full evaluation results, showing cumulative alignment error for all tested sequences in the TUM-monoVO dataset. Each square corresponds to the color-coded alignment error, as defined in \cite{engel2016photometrically}. We run each of the tested sequence (horizontal axis) 5 times each (vertical axis).}
      \label{fig:full_evaluation_res}
\end{figure*}

\subsection{Dataset Evaluation}
The first set of experiments focuses on evaluating H-SLAM performance on the TUM monoVO dataset \cite{engel2016photometrically}, which originally comprises 50 sequences with photometric calibration, recorded across diverse environments, including both indoor and outdoor settings, as well as transitions from indoor to outdoor environments. However, due to the limited configuration of photometric calibration, we utilized only 28 sequences captured by a narrow FOV camera. This dataset demonstrates the impact of our proposed photometric approach in response to changes in lighting conditions. 

H-SLAM undergoes a comparative analysis with the following SLAM systems:
\begin{itemize}
\item Direct Sparse Odometry (DSO) \cite{engel2017direct}: DSO employs a fully direct probabilistic model, minimizing photometric errors while jointly optimizing model parameters without the need for keypoint detectors or descriptors.
\item ORB-SLAM3 \cite{campos2021orb}: ORB-SLAM3 is a feature-based SLAM system designed to support wide baseline loop closing and relocalization, with the added feature of full automatic initialization.
\item Hybrid Sparse Monocular Visual Odometry (HSO) \cite{luo2022hybrid}: HSO utilizes direct image alignment with adaptive mode selection and employs ratio factors for image photometric description, enhancing robustness against substantial changes in image intensity and motion blur. 
\end{itemize}

All experiments were performed on an Intel i7-9750H CPU with 16 GB RAM, without GPU parallelization. Real-time execution was enforced, and the GUI was disabled to enhance performance. To assess our photometrically calibrated H-SLAM, we utilized the TUM calibration sequence with the OSPC to derive the camera response function, vignetting, and exposure time for each frame. 

\subsection{Evaluation in diverse operational settings}
Our approach to sequential photometric calibration requires the availability of exposure values from frame meta-data. This information is not provided in most publicly available SLAM benchmarking datasets. Hence, to further evaluate the performance of H-SLAM we conducted a second set of experiments on campus at the American University of Beirut.

The experiments were conducted in an outdoor areas spanning approximately 30mx40m, and ground truth GNSS trajectory data were obtained using a Real-Time Kinematic (RTK) GPS setup, ensuring centimeter-level accuracy for positional data. This RTK-GPS system utilizes both a stationary and a rover GNSS receiver, employing differential corrections to achieve its high precision. The ground truth trajectory serves as a benchmark for evaluating the trajectories generated by each of the tested SLAM systems.

HSO, ORB-SLAM, DSO, and H-SLAM underwent testing in the recorded environment to assess their performance. Recognizing the probabilistic nature of SLAM systems, each system was run five times, and the best-performing result was considered for analysis. This approach ensures a comprehensive evaluation, capturing the variability inherent in SLAM system performance.
\begin{figure}[ht]
  \includegraphics[width=\linewidth, height=70mm, margin=0]{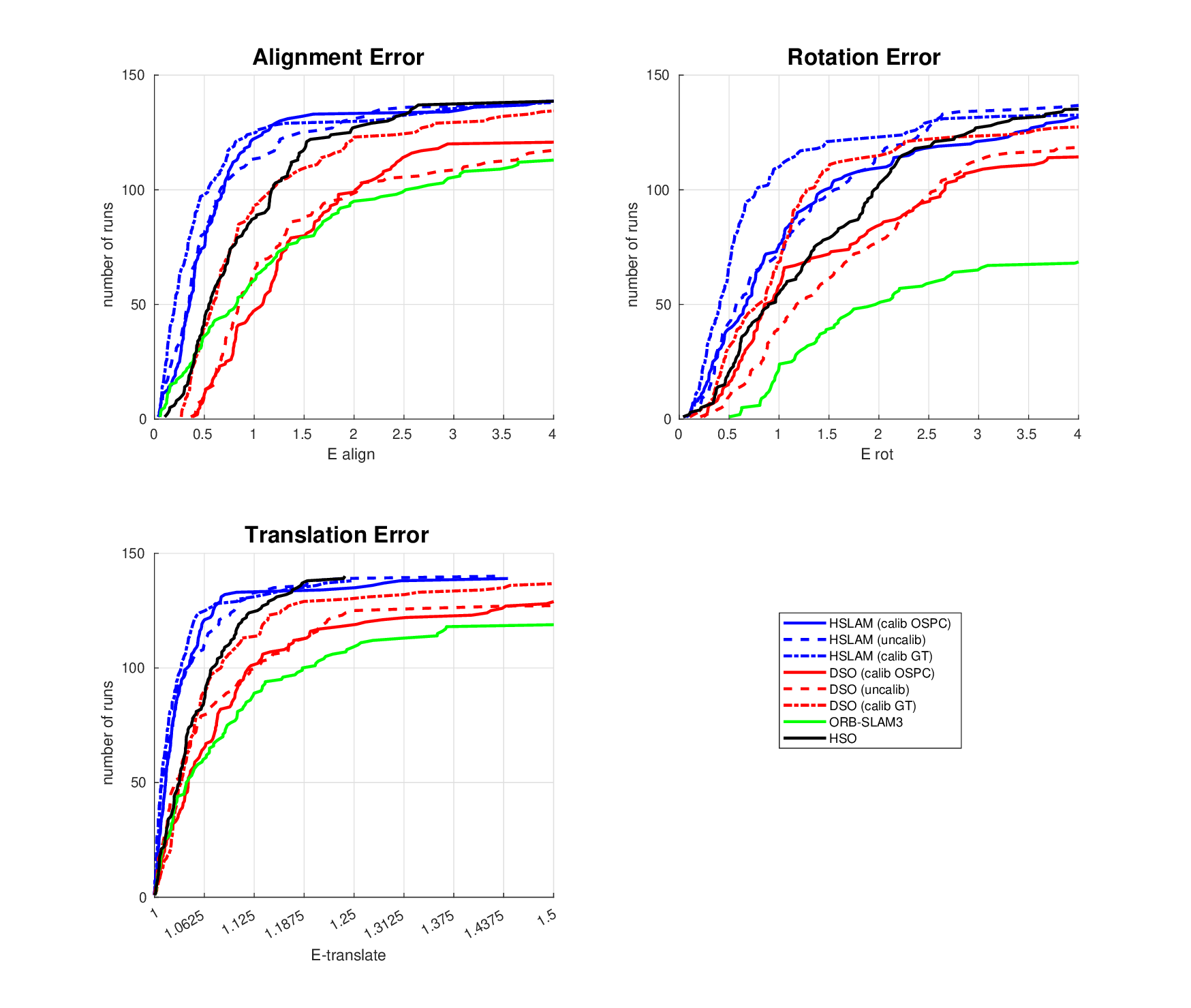}
  \caption{Accumulated alignment error, rotational drift error and translational error as defined in \cite{engel2016photometrically} for each system over all runs}
  \label{fig:cumulative_err}
\end{figure}

\subsection{Live real-time evaluation}
The third set of experiments took place indoors on Level 4 of the Irany Oxy Engineering Complex (IOEC) at the American University of Beirut (Fig. \ref{fig:qual_exp_res}). These experiments aimed to qualitatively assess the real-time performance of the SLAM systems. Utilizing the ROS architecture, all systems were simultaneously executed on the same machine. A single image frame captured by the camera was broadcasted across the ROS network, ensuring that each system received identical real-time input data. Key observations were focused on evaluating the quality of live performance and identifying any instances of failure among the SLAM systems. 
In all experiments, H-SLAM's loop closure feature was disabled to ensure a fair comparison with DSO, preventing any potential bias that would skew the results in favor of H-SLAM.

\section{Results and discussion}
\subsection{Dataset observations}
The detailed evaluation outcomes for each sequence is present in Fig. \ref{fig:full_evaluation_res}. ORB-SLAM3 encounters challenges in Sequences 35 to 40, all of which involve indoor settings characterized by featureless hallways and plain classrooms. Conversely, the algorithm demonstrates strong performance in outdoor sequences (45-48 and 29), where the environment was rich in distinctive features.
Similarly, in the absence of photometric calibration, DSO faces difficulties mapping indoor sequences that lack distinctive features. However, its performance improves in such scenarios when combined with OSPC, reaching its optimal capability when utilizing the provided DSO calibration. Additionally, DSO encounters challenges in mapping Sequences 26 and 27, both of which involve staircases. It is important to highlight that the DSO variant incorporating loop closure, LDSO, exhibited poor performance along all sequences and thus was not reported in the results. This poor performancee can be attributed to the TUM-VO dataset's inherent nature, characterized by huge closed loops which makes it difficult to associate the current data with the historical data. On the other hand, both HSO and H-SLAM consistently exhibit good performance across the sequence variations. Both algorithms effectively handle mapping tasks in indoor featureless scenarios and outdoor settings which proves the robustness of the proposed hybrid systems.
\begin{figure*}[htbp]
  \includegraphics[width=\linewidth, height=0.45\linewidth]{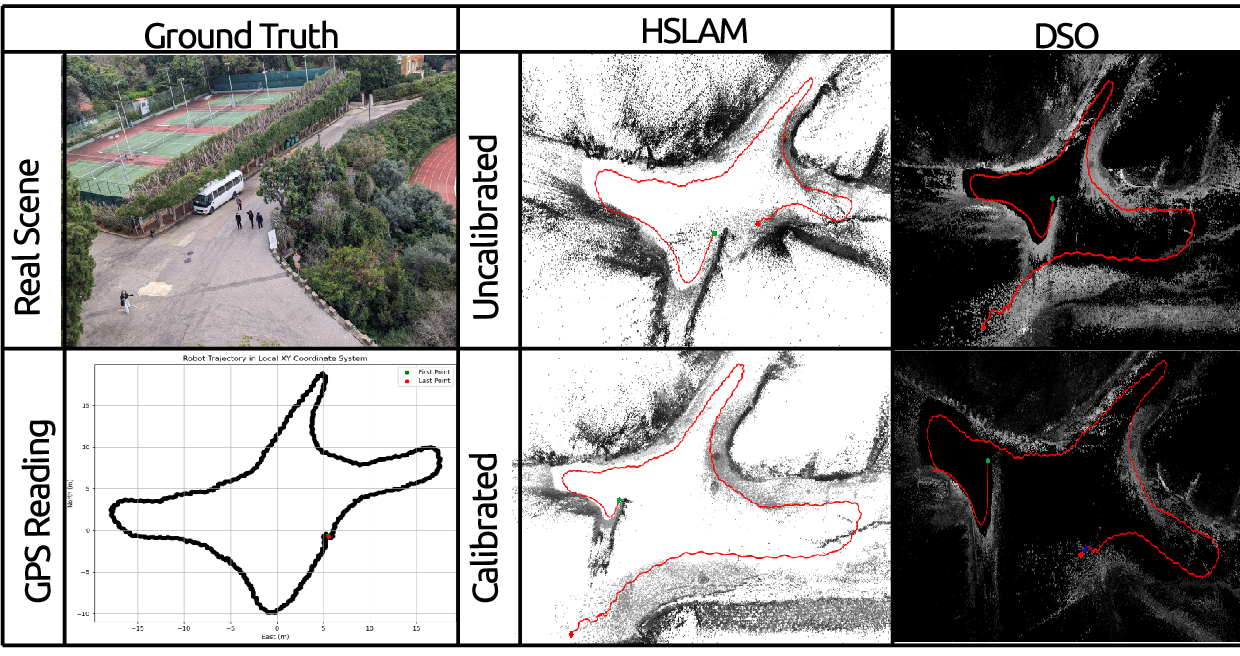}
  \caption{Tabulated results of the live experiment conducted on campus at the American University of Beirut. The table shows the trajectories of HSLAM and DSO under two calibration modes alongside the "ground truth" obtained from RTK-GPS}
  \label{fig:live_exp}
\end{figure*}
To facilitate a comprehensive comparison of the performance across the four algorithms, we calculated the cumulative alignment, rotation, and translation errors over the course of multiple runs, as defined in \cite{engel2017direct}. The resulting outcomes are presented in Fig. \ref{fig:cumulative_err}.
The graphs demonstrate H-SLAM consistently outperforming other algorithms by most metrics, regardless of the applied photometric calibration. Remarkably, even a photometrically uncalibrated system utilizing H-SLAM exhibits better performance than photometrically calibrated DSO, online photometrically calibrated HSO, and ORB-SLAM3, underscoring the robustness of the proposed system. HSO performance closely follows in our evaluation, exhibiting approximately similar performance to the calibrated DSO when considering ground truth photometric values.

Conversely, the graphs reveal the sensitivity of DSO to the photometric component. An uncalibrated DSO system proves to be the least performing, with gradual performance improvements when OSPC calibration is applied, reaching peak performance when ground truth photometric values are utilized. 
section{Outdoor Experiments Observations}
When applied to the dataset captured at AUB, both HSO and ORB-SLAM did not perform well. While both systems initialized properly, they struggled to maintain tracking as the scene transitioned into frames featuring the asphalt ground with sparse features. In contrast, DSO and H-SLAM, with their direct modules, demonstrated robustness in handling frame sequences with sparse features. The results produced by DSO and H-SLAM are presented in Table \ref{fig:live_exp}. In the absence of photometric calibration, both DSO and H-SLAM exhibit significant error drift. However, upon the application of photometric calibration (OSPC) both DSO and H-SLAM showcased a notable reduction in geometric and photometric residuals, thereby mitigating the overall drift error.

A photometrically calibrated H-SLAM demonstrates more accurate pose estimation compared to its uncalibrated counterpart, resulting in a final trajectory that is closer to the groundtruth GNSS data 

\subsection{Live Real-Time Experiments}
\begin{figure}[htbp]
  \includegraphics[width=\linewidth]{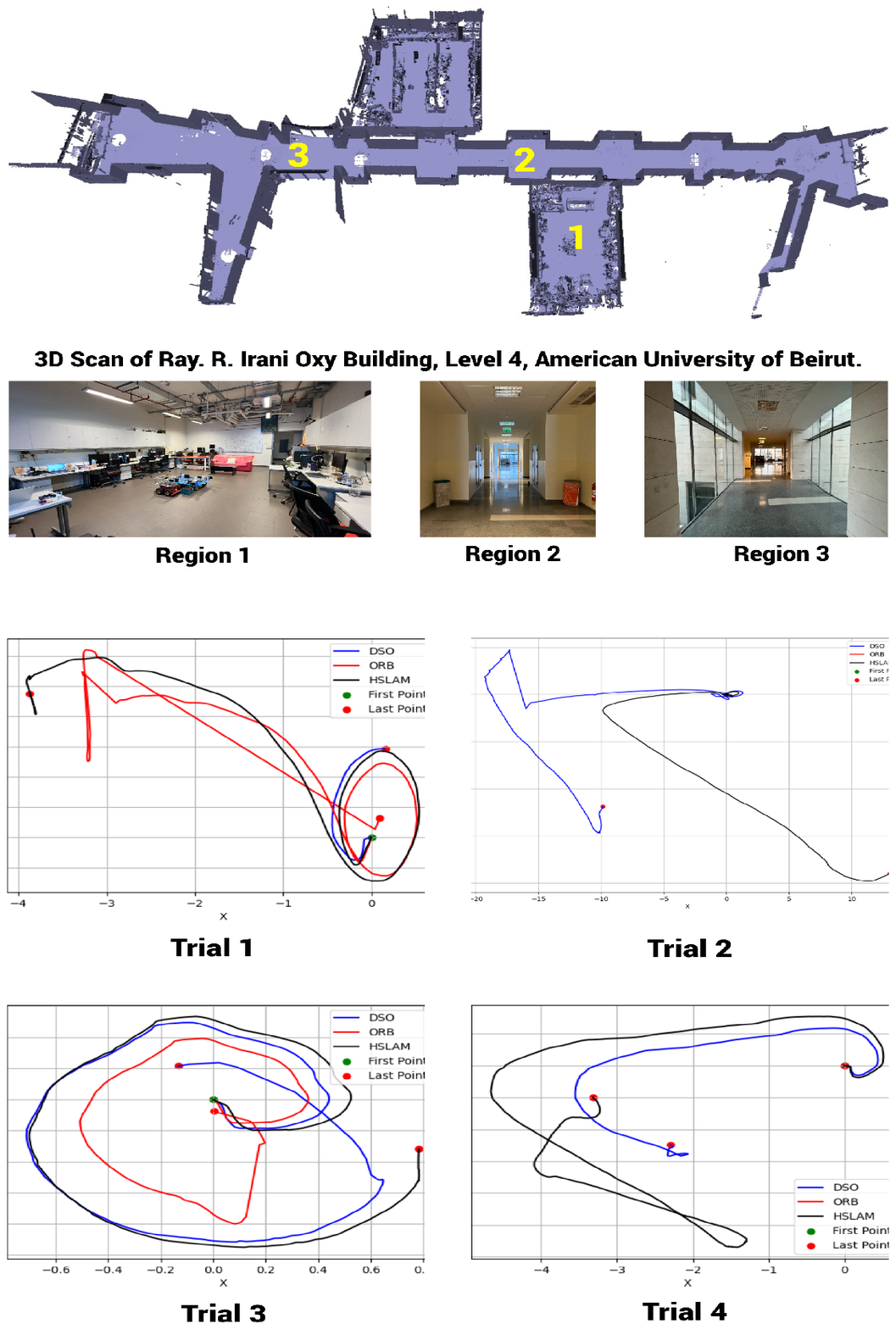}
  \caption{Trajectory generated by H-SLAM, ORB-SLAM and DSO during the live experiments, above which is a 3D scan and images of the area used in testing }
  \label{fig:qual_exp_res}
\end{figure}

The results of the live real-time experiments are depicted in Fig. \ref{fig:qual_exp_res}. 
In Trial 1, conducted in Region 1, the trajectory formed a loop within the lab, extending to the lab door leading to the corridor (Region 2). DSO initially tracks the trajectory but loses track when encountering a bright glare from the lab window. ORB loses tracking at the door and exhibits erratic behavior, and H-SLAM is stopped when ORB loses track. Trial 2 commences from Region 1 through the door and into the corridor (Region 2). ORB fails to initialize, while DSO initially runs but loses track at the corridor door before reinitializing. However, H-SLAM encounters no issues throughout the trial. Trial 3 involves a simple loop within Region 1, showcasing trajectories drawn by the systems in a confined space. ORB and DSO produce trajectories deviating from the actual motion, with DSO briefly losing track and reinitializing, while H-SLAM exhibits the most accurate trajectory. Trial 4, conducted in Region 2 with a turn at the glass walls (Region 3), sees ORB failing to initialize. DSO successfully tracks until encountering the turn at the glass walls, where it loses track entirely. H-SLAM operates without issues during this trial.

Overall, H-SLAM  demonstrates greater stability and lower sensitivity to light changes compared to the other two systems, exhibiting a lower rate of tracking loss and re-initialization.


\section{Conclusion}
This paper presented an enhanced performance of V-SLAM systems in real-world applications, integrating OSPC for improved photometric calibration and feature extraction within the H-SLAM framework. Experiments on two datasets demonstrated superior performance for our proposed calibrated H-SLAM versus state of the art in V-SLAM systems.
We are currently working on integrating Inertial Measurement Units (IMU) into H-SLAM system to provide more robust solutions for navigation and mapping in complex environments.

\section*{Acknowledgment}
This work was supported by the DIDYMOS-XR Horizon Europe project (grant number 101092875–DIDYMOS-XR, www.didymos-xr.eu), the Dean's Office and the Vertically Integrated Projects Program at Maroun Semaan Faculty of Engineering and Architecture at AUB.

\bibliographystyle{IEEEtran}
\bibliography{main.bib}

\begin{thebibliography}{10}
\providecommand{\url}[1]{#1}
\csname url@samestyle\endcsname
\providecommand{\newblock}{\relax}
\providecommand{\bibinfo}[2]{#2}
\providecommand{\BIBentrySTDinterwordspacing}{\spaceskip=0pt\relax}
\providecommand{\BIBentryALTinterwordstretchfactor}{4}
\providecommand{\BIBentryALTinterwordspacing}{\spaceskip=\fontdimen2\font plus
\BIBentryALTinterwordstretchfactor\fontdimen3\font minus \fontdimen4\font\relax}
\providecommand{\BIBforeignlanguage}[2]{{%
\expandafter\ifx\csname l@#1\endcsname\relax
\typeout{** WARNING: IEEEtran.bst: No hyphenation pattern has been}%
\typeout{** loaded for the language `#1'. Using the pattern for}%
\typeout{** the default language instead.}%
\else
\language=\csname l@#1\endcsname
\fi
#2}}
\providecommand{\BIBdecl}{\relax}
\BIBdecl

\bibitem{engel2017direct}
J.~Engel, V.~Koltun, and D.~Cremers, ``Direct sparse odometry,'' \emph{IEEE transactions on pattern analysis and machine intelligence}, vol.~40, no.~3, pp. 611--625, 2017.

\bibitem{bergmann2017online}
P.~Bergmann, R.~Wang, and D.~Cremers, ``Online photometric calibration of auto exposure video for realtime visual odometry and slam,'' \emph{IEEE Robotics and Automation Letters}, vol.~3, no.~2, pp. 627--634, 2017.

\bibitem{mur2015orb}
R.~Mur-Artal, J.~M.~M. Montiel, and J.~D. Tardos, ``Orb-slam: a versatile and accurate monocular slam system,'' \emph{IEEE transactions on robotics}, vol.~31, no.~5, pp. 1147--1163, 2015.

\bibitem{grossberg2003determining}
M.~D. Grossberg and S.~K. Nayar, ``Determining the camera response from images: What is knowable?'' \emph{IEEE Transactions on pattern analysis and machine intelligence}, vol.~25, no.~11, pp. 1455--1467, 2003.

\bibitem{haidar2023ospc}
J.~Haidar, D.~Khalil, and D.~Asmar, ``Ospc: Online sequential photometric calibration,'' \emph{arXiv preprint arXiv:2305.17673}, 2023.

\bibitem{younes2023h}
G.~Younes, D.~Khalil, J.~Zelek, and D.~Asmar, ``H-slam: Hybrid direct-indirect visual slam,'' \emph{arXiv preprint arXiv:2306.07363}, 2023.

\bibitem{geiger2011stereoscan}
A.~Geiger, J.~Ziegler, and C.~Stiller, ``Stereoscan: Dense 3d reconstruction in real-time,'' in \emph{2011 IEEE intelligent vehicles symposium (IV)}.\hskip 1em plus 0.5em minus 0.4em\relax Ieee, 2011, pp. 963--968.

\bibitem{klein2007parallel}
G.~Klein and D.~Murray, ``Parallel tracking and mapping for small ar workspaces,'' in \emph{2007 6th IEEE and ACM international symposium on mixed and augmented reality}.\hskip 1em plus 0.5em minus 0.4em\relax IEEE, 2007, pp. 225--234.

\bibitem{pire2015stereo}
T.~Pire, T.~Fischer, J.~Civera, P.~De~Crist{\'o}foris, and J.~J. Berlles, ``Stereo parallel tracking and mapping for robot localization,'' in \emph{2015 IEEE/RSJ international conference on intelligent robots and systems (IROS)}.\hskip 1em plus 0.5em minus 0.4em\relax IEEE, 2015, pp. 1373--1378.

\bibitem{engel2013semi}
J.~Engel, J.~Sturm, and D.~Cremers, ``Semi-dense visual odometry for a monocular camera,'' in \emph{Proceedings of the IEEE international conference on computer vision}, 2013, pp. 1449--1456.

\bibitem{newcombe2011dtam}
R.~A. Newcombe, S.~J. Lovegrove, and A.~J. Davison, ``Dtam: Dense tracking and mapping in real-time,'' in \emph{2011 international conference on computer vision}.\hskip 1em plus 0.5em minus 0.4em\relax IEEE, 2011, pp. 2320--2327.

\bibitem{lee2018loosely}
S.~H. Lee and J.~Civera, ``Loosely-coupled semi-direct monocular slam,'' \emph{IEEE Robotics and Automation Letters}, vol.~4, no.~2, pp. 399--406, 2018.

\bibitem{forster2016svo}
C.~Forster, Z.~Zhang, M.~Gassner, M.~Werlberger, and D.~Scaramuzza, ``Svo: Semidirect visual odometry for monocular and multicamera systems,'' \emph{IEEE Transactions on Robotics}, vol.~33, no.~2, pp. 249--265, 2016.

\bibitem{kim2019autonomous}
P.~Kim, H.~Lee, and H.~J. Kim, ``Autonomous flight with robust visual odometry under dynamic lighting conditions,'' \emph{Autonomous Robots}, vol.~43, no.~6, pp. 1605--1622, 2019.

\bibitem{krombach2017combining}
N.~Krombach, D.~Droeschel, and S.~Behnke, ``Combining feature-based and direct methods for semi-dense real-time stereo visual odometry,'' in \emph{Intelligent Autonomous Systems 14: Proceedings of the 14th International Conference IAS-14 14}.\hskip 1em plus 0.5em minus 0.4em\relax Springer, 2017, pp. 855--868.

\bibitem{luo2022hybrid}
D.~Luo, Y.~Zhuang, and S.~Wang, ``Hybrid sparse monocular visual odometry with online photometric calibration,'' \emph{The International Journal of Robotics Research}, vol.~41, no. 11-12, pp. 993--1021, 2022.

\bibitem{younes2019unified}
G.~Younes, D.~Asmar, and J.~Zelek, ``A unified formulation for visual odometry,'' in \emph{2019 IEEE/RSJ International Conference on Intelligent Robots and Systems (IROS)}.\hskip 1em plus 0.5em minus 0.4em\relax IEEE, 2019, pp. 6237--6244.

\bibitem{civera2008inverse}
J.~Civera, A.~J. Davison, and J.~M. Montiel, ``Inverse depth parametrization for monocular slam,'' \emph{IEEE transactions on robotics}, vol.~24, no.~5, pp. 932--945, 2008.

\bibitem{mann1995being}
S.~Mann and R.~Picard, ``On being” undigital” with digital cameras: Extending dynamic range by combining differently exposed pictures, roceedings of is\&t 48th annual conference society for imaging science and technology annual conference,'' 1995.

\bibitem{mitsunaga1999radiometric}
T.~Mitsunaga and S.~K. Nayar, ``Radiometric self calibration,'' in \emph{Proceedings. 1999 IEEE computer society conference on computer vision and pattern recognition (Cat. No PR00149)}, vol.~1.\hskip 1em plus 0.5em minus 0.4em\relax IEEE, 1999, pp. 374--380.

\bibitem{grossberg2003space}
M.~D. Grossberg and S.~K. Nayar, ``What is the space of camera response functions?'' in \emph{2003 IEEE Computer Society Conference on Computer Vision and Pattern Recognition, 2003. Proceedings.}, vol.~2.\hskip 1em plus 0.5em minus 0.4em\relax IEEE, 2003, pp. II--602.

\bibitem{lin2004radiometric}
J.~Gu and H.-Y. Shum, ``Radiometric calibration from a single image,'' in \emph{Proceedings of the 2004 IEEE Computer Society Conference on Computer Vision and Pattern Recognition, 2004. CVPR 2004.}, vol.~2.\hskip 1em plus 0.5em minus 0.4em\relax IEEE, 2004, pp. II--II.

\bibitem{lin2005determining}
S.~Lin and L.~Zhang, ``Determining the radiometric response function from a single grayscale image,'' in \emph{2005 IEEE Computer Society Conference on Computer Vision and Pattern Recognition (CVPR'05)}, vol.~2.\hskip 1em plus 0.5em minus 0.4em\relax IEEE, 2005, pp. 66--73.

\bibitem{debevec2023recovering}
P.~E. Debevec and J.~Malik, ``Recovering high dynamic range radiance maps from photographs,'' in \emph{Seminal Graphics Papers: Pushing the Boundaries, Volume 2}, 2023, pp. 643--652.

\bibitem{engel2016photometrically}
J.~Engel, V.~Usenko, and D.~Cremers, ``A photometrically calibrated benchmark for monocular visual odometry,'' \emph{arXiv preprint arXiv:1607.02555}, 2016.

\bibitem{zheng2008single}
Y.~Zheng, S.~Lin, C.~Kambhamettu, J.~Yu, and S.~B. Kang, ``Single-image vignetting correction,'' \emph{IEEE transactions on pattern analysis and machine intelligence}, vol.~31, no.~12, pp. 2243--2256, 2008.

\bibitem{kim2008robust}
S.~J. Kim and M.~Pollefeys, ``Robust radiometric calibration and vignetting correction,'' \emph{IEEE transactions on pattern analysis and machine intelligence}, vol.~30, no.~4, pp. 562--576, 2008.

\bibitem{mo2019ambiguity}
Z.~Mo, B.~Shi, S.-K. Yeung, and Y.~Matsushita, ``Ambiguity-free radiometric calibration for internet photo collections,'' \emph{IEEE Transactions on Pattern Analysis and Machine Intelligence}, vol.~42, no.~7, pp. 1670--1684, 2019.

\bibitem{liu2021optical}
Q.~Liu, Z.~Wang, and H.~Wang, ``Optical flow monocular visual-inertial odometry with online photometric calibration,'' in \emph{Journal of Physics: Conference Series}, vol. 1828, no.~1.\hskip 1em plus 0.5em minus 0.4em\relax IOP Publishing, 2021, p. 012163.

\bibitem{campos2021orb}
C.~Campos, R.~Elvira, J.~J.~G. Rodr{\'\i}guez, J.~M. Montiel, and J.~D. Tard{\'o}s, ``Orb-slam3: An accurate open-source library for visual, visual--inertial, and multimap slam,'' \emph{IEEE Transactions on Robotics}, vol.~37, no.~6, pp. 1874--1890, 2021.

\end{thebibliography}
\end{document}